\documentclass[11pt]{article}

\usepackage[margin=1in]{geometry}
\usepackage{times}
\usepackage{microtype}
\usepackage{graphicx}
\usepackage{booktabs}
\usepackage{multirow}
\usepackage{amsmath,amssymb,amsfonts}
\usepackage{mathtools}
\usepackage{enumitem}
\usepackage{xcolor}
\usepackage{hyperref}
\usepackage{url}
\usepackage{caption}
\usepackage{subcaption}
\usepackage{array}
\usepackage{placeins}
\usepackage{tabularx}
\usepackage{siunitx}
\usepackage[numbers,sort&compress]{natbib}
\usepackage{xspace}

\graphicspath{{figures/}}
\hypersetup{
  colorlinks=true,
  linkcolor=blue!50!black,
  citecolor=blue!50!black,
  urlcolor=blue!50!black,
}

\newcommand{\qift}{Qift\xspace}
\newcommand{\nz}{\textsc{MNZ}\xspace}
\newcommand{\potwide}{\textsc{PoT-MNZ}\xspace}
\newcommand{\qiftnz}{Qift-\textsc{MNZ}\xspace}
\newcommand{\qiftpot}{Qift-\textsc{PoT-MNZ}\xspace}
\newcommand{\symint}{\textsc{SYM-INT}}
\newcommand{\wasym}{\textsc{W-ASYM}\xspace}
\newcommand{\lloyd}{\textsc{Lloyd}}
\newcommand{\nf}{\textsc{NF2}}
\newcommand{\farnz}{\textsc{Far-MNZ}\xspace}

\newcommand{\figmaybe}[3]{%
\IfFileExists{figures/#1}{\includegraphics[width=#2]{#1}}{\fbox{\parbox{#2}{\centering Figure file not found: #1\\#3}}}}

\title{Qift: Shift-Friendly No-Zero W2 Post-Training Quantization for Rotated W2A4/KV4 LLM Inference}

\author{
Chi-Wei Huang\\
National Cheng Kung University\\
\texttt{m56121041@gs.ncku.edu.tw}
\and
Chia-Chi Tsai\\
National Cheng Kung University\\
\texttt{cctsai@gs.ncku.edu.tw}
}

\date{}

\begin{document}
\maketitle

\begin{abstract}
Two-bit weight quantization is attractive for memory-efficient large language model (LLM) inference, but the standard W2 level set $\{-2,-1,0,+1\}$ often collapses under aggressive W2A4/KV4 settings. We study the scalar level-set geometry of two-bit weights in a Hadamard-rotated quantization pipeline. Conventional asymmetric W2 substantially improves over the standard level set, indicating that W2A4 failure is not only a bit-width problem but also a reconstruction-level problem. Across all 224 linear modules in each of LLaMA-2-7B and LLaMA-3.1-8B, we show that pretrained weights are already nearly zero-centered while Hadamard rotation primarily Gaussianizes their standardized shape: excess kurtosis and Q--Q error drop by orders of magnitude and skewness also decreases substantially, while the per-channel mean stays close to zero relative to the per-channel standard deviation.

Based on this approximate zero-centered Gaussian-like source model, we propose \qift, a fixed no-zero W2 level set for rotated W2A4/KV4 inference. The main level set is $\{\pm0.5,\pm1.5\}$, equivalently $\{\pm1,\pm3\}$ under a half-scale reparameterization; a power-of-two variant uses $\{\pm1,\pm4\}$ for sign-and-shift decoded weight application. \qift redesigns the fixed two-bit code-to-level mapping and is training-free, learned-codebook-free, group-grid-free, and zero-point-free, retaining the standard per-channel scale. A scale-invariant ratio analysis identifies an effective inner/outer centroid ratio range of $0.25$--$0.33$, explaining why mirror no-zero (\nz), Lloyd, NF2, and \potwide perform well while $\{\pm1,\pm2\}$ does not.

Experiments on LLaMA-2-7B and LLaMA-3.1-8B show that the proposed no-zero level sets consistently improve pure W2A4 perplexity, L-layer mixed W2/W4 perplexity, downstream accuracy, and GPTQ residual behavior over the standard W2 level set; as the ratio analysis indicates, this improvement requires an appropriate inner/outer ratio rather than the mere absence of a zero level, and a no-zero set with too large a ratio such as $\{\pm1,\pm2\}$ does not improve over the standard level set. At $L{=}16$ mixed precision, no-zero level sets substantially narrow the gap to W3A4 while keeping half of the transformer layers at two-bit precision. Fixed no-zero scalar level sets provide a simple, source-aware, and hardware-aligned alternative to more complex learned W2 codebooks for rotated W2A4/KV4 inference.
\end{abstract}

\section{Introduction}

Decode-phase large language model (LLM) inference is highly memory-bound: during autoregressive decoding, throughput is limited primarily by the cost of moving weights and the key--value (KV) cache from memory rather than by arithmetic. Weight and KV-cache bit-width therefore directly bound achievable throughput on constrained hardware, which makes low-bit quantization a central lever for efficient inference. Rotation-based post-training quantization (PTQ) has made W4A4/KV4 inference increasingly practical and close to lossless, and W3A4 is an aggressive but workable operating point. The more extreme W2A4/KV4 regime---an $8\times$ weight compression target relative to FP16---remains far less explored and is highly sensitive to quantizer design. In particular, its scalar reconstruction level set has largely been inherited from the standard symmetric integer quantizer rather than designed for the rotated weight distribution.

Existing low-bit LLM quantization methods combine several complementary ingredients. Smoothing shifts activation outlier difficulty into weights, rotations diffuse channel-wise outlier energy, calibration and compensation methods such as GPTQ and GPTAQ reduce the remaining discretization error, mixed precision protects sensitive layers with additional bits, and weight-only quantization keeps activations in high precision to avoid compounding weight and activation quantization error. These techniques are crucial for practical low-bit inference and also form the foundation of the rotated W2A4/KV4 setting studied here.

However, once weights are reduced to two bits, a different error source becomes first-order. With only four reconstruction levels, the dense central bulk of the rotated weight distribution is no longer represented almost for free. Outlier handling and compensation are necessary but no longer sufficient: they operate with respect to a chosen W2 level set, but they do not determine whether those four levels represent the dense rotated bulk well. Thus, to push beyond practical W4A4/KV4 toward the more extreme W2A4/KV4 regime, this work treats the W2 reconstruction level set itself as a first-class design variable.

This work focuses on the scalar level set used by two-bit weight quantization inside a rotated LLM quantization pipeline. In many symmetric integer quantizers, a \(b\)-bit weight is represented by a signed integer code
\begin{equation}
    q = \mathrm{clip}\!\left(
        \left\lfloor \frac{w}{s} \right\rceil,
        -2^{b-1},
        2^{b-1}-1
    \right),
    \qquad
    \hat{w}=s q,
\end{equation}
where \(s\) is a per-output-channel scale. For \(b=2\), this gives the standard W2 reconstruction level set
\begin{equation}
    \mathcal{G}_{\mathrm{sym}} = \{-2,-1,0,+1\}.
\end{equation}
Equivalently, the four reconstruction levels are \(s\mathcal{G}_{\mathrm{sym}}\). In practice, \(s\) may be chosen by max-range or clipping-based MSE search, but the scalar reconstruction geometry is still determined by \(\mathcal{G}_{\mathrm{sym}}\).

We use reconstruction level set for the four scalar values used by a W2 quantizer, and ``grid'' only as an informal synonym. In scalar-quantization terminology, a mid-tread quantizer includes zero as a reconstruction level, whereas a mid-rise quantizer places zero between two reconstruction levels. The standard signed-integer W2 grid, denoted \symint, is mid-tread-style: it is simple, but it spends one of only four reconstruction levels exactly at zero, which is not a neutral choice for a two-bit quantizer.

A direct indication that the grid, and not merely the bit-width, is responsible comes from conventional asymmetric W2 quantization: simply replacing the standard \symint\ grid with an asymmetric W2 quantizer substantially reduces pure-W2A4 perplexity (on LLaMA-2-7B with KV4 and GPTQ, from 53.849 to 33.533; on LLaMA-3.1-8B, from 3005.556 to 113.747). We treat this as evidence---rather than as the central result---that the placement of the four reconstruction levels is a dominant factor in W2A4 failure, and we use it only to motivate a principled, source-aware grid design.

From a quantization perspective, the tensor distribution determines how efficiently a small number of reconstruction levels can represent values. A compact and approximately symmetric source is easier to quantize because most values lie near the center and the quantizer can allocate levels to high-density regions. In contrast, skewed or heavy-tailed sources waste range on rare extremes and reduce effective precision for the dense central mass. Thus, Gaussian-like is not the final goal by itself; it is a useful reference for a centered, symmetric, low-outlier source.

This motivates the central question of this work: after Hadamard rotation, what scalar level set should a two-bit weight quantizer use? We show that rotated weights are better modeled as approximately zero-centered and more Gaussian-like in standardized shape. For such a source, a four-level scalar quantizer should place two inner centroids around zero and two outer centroids in the tails, instead of spending a centroid exactly at zero.

We call the proposed design \qift, short for quantization with shift-friendly no-zero W2 grids. The main practical level set is mirror no-zero (\nz),
\begin{equation}
    \mathcal{G}_{\mathrm{MNZ}} = \{-1.5,-0.5,+0.5,+1.5\},
\end{equation}
a uniform four-level mid-rise level set, equivalent to odd integer levels $\{\pm1,\pm3\}$ under a half-scale reparameterization. We also study a power-of-two variant, \potwide,
\begin{equation}
    \mathcal{G}_{\mathrm{pot}} = \{-4,-1,+1,+4\},
\end{equation}
which keeps the mirror no-zero structure but is not a uniform mid-rise grid; its power-of-two magnitudes support sign-and-shift decoded weight application. In both cases, the per-channel scale can be applied separately in the epilogue, as in standard quantized linear layers.

The proposed grids are deliberately minimal: they use the same four reconstruction levels globally, retain only the standard per-channel scale, and introduce no quantization-aware training, learned partitions, per-layer or group-wise grid assignment (we call this group-grid-free), learned codebooks, or asymmetric zero-points. Thus, \qift\ treats W2 quantization as an accuracy-aware level-set design problem while remaining a drop-in replacement for the standard W2 grid rather than a new learned quantizer.

\paragraph{Contributions.}
Our contributions are summarized as follows.
\begin{itemize}
    \item We introduce \qift as a modular reconstruction-level redesign for rotated W2A4/KV4 inference.
    It isolates the fixed W2 level set as the design intervention, keeping the surrounding rotation, scaling, and PTQ compensation pipeline unchanged and requiring no quantization-aware training, learned codebooks, group-wise grid assignment, or zero-point metadata.

    \item We propose mirror no-zero (\nz), a fixed no-zero W2 scalar level set for approximately zero-centered Gaussian-like rotated weights.
    \nz provides a simple integer approximation to the four-level Gaussian Lloyd-Max structure, together with a power-of-two \potwide variant for sign-and-shift decoded weight application.

    \item We validate \qift across two LLaMA models.
    The proposed level sets consistently improve over the standard W2 grid in pure W2A4, mixed W2/W4, and downstream tasks, while ablations show that the gain comes from an effective inner/outer centroid ratio rather than zero removal alone.
\end{itemize}

\section{Related Work}

We organize prior low-bit LLM quantization by which component is primarily changed. Some methods transform the tensor distribution seen by the quantizer; others improve calibration or error compensation after discretization; and a third group changes the quantizer, reconstruction levels, partitions, or codebooks themselves. Table~\ref{tab:taxonomy-related} summarizes this view and locates \qift\ within it.

\paragraph{Equivalent transformations.}
A major line of work improves quantization by applying mathematically equivalent transformations to weights and activations so that the resulting tensors are easier to quantize. SmoothQuant~\citep{smoothquant} migrates activation outlier difficulty into weights through channel-wise scaling, while rotation-based methods reduce outliers with orthogonal or learned transforms: QuaRot~\citep{quarot} uses Hadamard rotations for end-to-end W4A4/KV4 inference, SpinQuant~\citep{spinquant} learns the rotation matrices, and FlatQuant~\citep{flatquant} learns affine transforms that flatten weight and activation statistics. These methods make tensors more quantization-friendly but keep the scalar integer reconstruction levels fixed. \qift\ is complementary: it assumes such a rotated pipeline and redesigns the W2 reconstruction levels used after the transformation.

\paragraph{Calibration and error compensation.}
Another line of work uses calibration data to reduce the effect of discretization rather than changing the reconstruction levels. GPTQ~\citep{gptq} uses a Hessian approximation to compensate weight quantization error, and GPTAQ~\citep{gptaq} adds activation-aware asymmetric calibration. AWQ~\citep{awq} uses activation statistics to identify and protect salient weights, and OmniQuant~\citep{omniquant} learns equivalent transformations and clipping parameters for post-training quantization. These methods improve the calibration or compensation procedure; \qift\ instead redesigns the fixed W2 reconstruction level set and can be combined with such pipelines, as in our GPTQ/GPTAQ experiments.

\paragraph{Quantizer and level-set design.}
A third line changes the quantizer itself---its reconstruction levels, partitions, or codebooks. RCP~\citep{rcp} is the closest prior work, as it also targets W2A4/KV4: it integrates rotation, clipping, and a learnable non-uniform W2 quantizer trained with quantization-aware training (QAT). NF4~\citep{qlora} uses a normal-distribution motivation for a 4-bit datatype, conceptually close to our Gaussian Lloyd-Max~\citep{lloyd,max} and NF2 references. QuIP\#~\citep{quip} and AQLM~\citep{aqlm} move beyond scalar quantization with lattice or additive vector codebooks, and LeanQuant~\citep{leanquant} learns loss-error-aware adaptive grids. \qift\ takes a different point in this design space: it keeps the quantizer scalar, fixed, training-free, and zero-point-free, and redesigns the four W2 reconstruction levels themselves. The contrast with RCP is direct---RCP learns non-uniform W2 partitions through QAT, whereas \qift\ uses a fixed, source-aware, post-training no-zero scalar level set.

\paragraph{Weight-only compression versus W2A4/KV4 inference.}
Many extreme low-bit methods, including QuIP\#~\citep{quip} and AQLM~\citep{aqlm}, primarily target weight-only compression, where the main benefit is reduced parameter storage and weight memory traffic. In contrast, this work studies a rotated W2A4/KV4 inference setting, where the W2 levels must interact with four-bit activations and KV-cache quantization. This makes the scalar W2 level-set design more constrained than in weight-only compression. The proposed no-zero level sets keep the quantizer fixed and scalar while avoiding learned codebook lookup and asymmetric zero-point metadata.

\begin{table}[t]
\centering
\caption{Taxonomy of related LLM quantization techniques and the position of \qift. The table groups methods by which component they primarily change; it is not a cross-paper accuracy leaderboard.}
\label{tab:taxonomy-related}
\small
\setlength{\tabcolsep}{4pt}
\renewcommand{\arraystretch}{1.2}
\begin{tabularx}{\linewidth}{%
  >{\raggedright\arraybackslash}p{2.4cm}%
  >{\raggedright\arraybackslash\hsize=.85\hsize}X%
  >{\raggedright\arraybackslash\hsize=1.25\hsize}X%
  >{\raggedright\arraybackslash\hsize=.90\hsize}X}
\toprule
Category & Key idea & Representative methods & Relation to \qift \\
\midrule
Equivalent transformation
& Transform tensors to reduce outliers
& SmoothQuant~\citep{smoothquant}, QuaRot~\citep{quarot}, SpinQuant~\citep{spinquant}, FlatQuant~\citep{flatquant}
& Complementary; \qift\ changes W2 reconstruction levels after rotation \\

Calibration / compensation
& Use calibration data to reduce quantization error
& GPTQ~\citep{gptq}, GPTAQ~\citep{gptaq}, AWQ~\citep{awq}, OmniQuant~\citep{omniquant}
& Orthogonal; \qift\ reuses the same PTQ compensation pipeline \\

Quantizer / level-set design
& Change reconstruction levels, partitions, or codebooks
& RCP~\citep{rcp}, NF4~\citep{qlora}, QuIP\#~\citep{quip}, AQLM~\citep{aqlm}, LeanQuant~\citep{leanquant}
& Closest category; \qift\ uses fixed no-zero scalar W2 levels \\

Inference setting
& Weight-only versus weight--activation--KV quantization
& QuIP\#~\citep{quip}, AQLM~\citep{aqlm}, QuaRot~\citep{quarot}, RCP~\citep{rcp}
& \qift\ targets rotated W2A4/KV4 rather than weight-only compression \\
\bottomrule
\end{tabularx}
\end{table}

\begin{table}[t]
\centering
\caption{Design-level positioning of W2 quantization choices.
The table compares decoding and metadata properties rather than reporting
cross-paper accuracy. CB-free = no learned codebook; ZP-free = no zero-point;
Group-free = reconstruction levels, learned centroids, or lookup tables are not
assigned per weight group (standard per-channel scaling is still used).}
\label{tab:design-positioning}
\footnotesize
\setlength{\tabcolsep}{4pt}
\renewcommand{\arraystretch}{1.1}
\begin{tabularx}{\linewidth}{>{\raggedright\arraybackslash}p{2.6cm}ccccc>{\raggedright\arraybackslash}X}
\toprule
Method / level set & QAT-free & Tuning-free & CB-free & ZP-free & Group-free & Apply \\
\midrule
RCP-style learned W2~\citep{rcp} & No & No & No & mixed & learned & learned non-uniform \\
Weight-only codebooks~\citep{quip,aqlm} & Yes & mixed & No & usually & No & lookup / vector codebook \\
Conventional \wasym & Yes & Yes & Yes & No & Yes & zero-point subtract \\
Standard \symint{} W2 & Yes & Yes & Yes & Yes & Yes & simple but mismatched \\
Lloyd~\citep{lloyd,max} / NF2~\citep{qlora} & Yes & Yes & fixed table & Yes & Yes & irregular scalar levels \\
\qiftnz & Yes & Yes & Yes & Yes & Yes & small signed integer levels \\
\qiftpot & Yes & Yes & Yes & Yes & Yes & power-of-two signed levels \\
\bottomrule
\end{tabularx}
\end{table}

Overall, existing methods mainly improve the tensor distribution, the
calibration or compensation procedure, or the expressiveness of the quantizer.
\qift\ focuses on a smaller but underexplored design variable: the fixed
four-level W2 reconstruction level set used inside a rotated W2A4/KV4 PTQ
pipeline. Table~\ref{tab:design-positioning} compares these methods by their
decoding and metadata properties---whether they are codebook-free,
zero-point-free, and group-free---and places \qift\ at the fixed, scalar,
no-zero corner of this space.

\section{\qift: No-Zero W2 Reconstruction-Level Design}

\subsection{Overview and Base Pipeline}
\begin{figure}[t]
\centering
\figmaybe{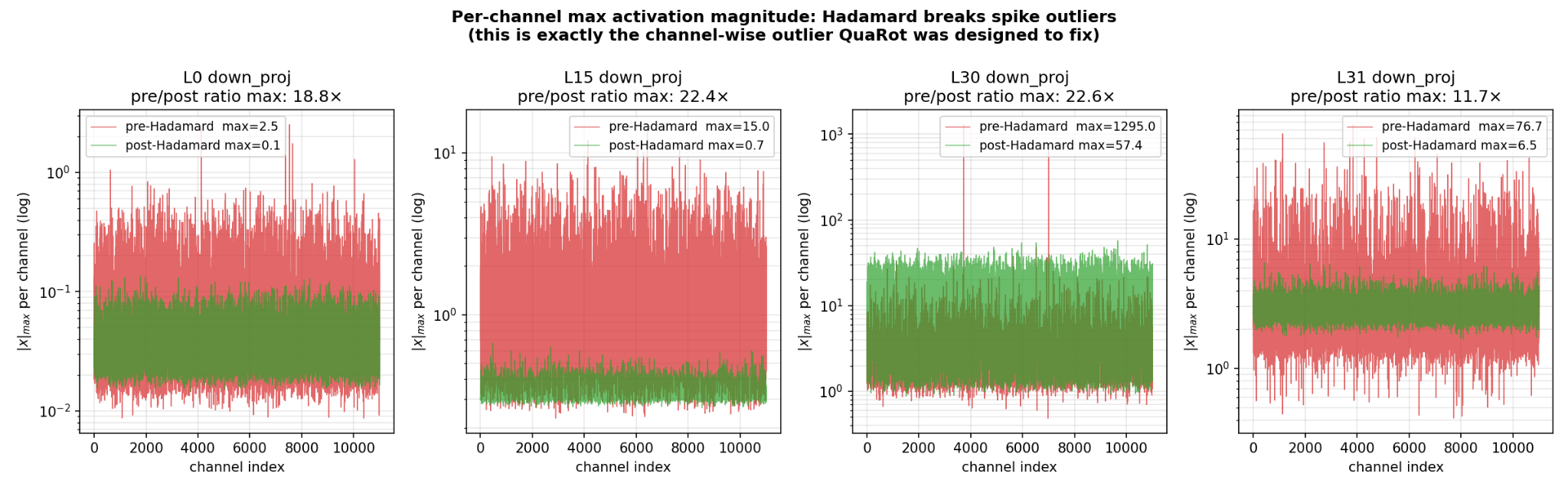}{0.9\linewidth}{Per-channel activation max.}
\caption{Channel-wise activation outlier reduction after Hadamard rotation for selected LLaMA-2-7B \texttt{down\_proj} inputs. We report the maximum activation magnitude for each channel, rather than the single global maximum over the entire activation tensor. The largest channel-wise spike in layer 30 drops from 1295.0 to 57.4, and the plotted layers show roughly 11.7--22.6$\times$ reduction in dominant channel-wise spikes. This activation-side outlier diffusion motivates the fixed rotated W2A4/KV4 pipeline used by \qift; the rest of this section studies the resulting weight-side reconstruction-level design.}
\label{fig:activation-max}
\end{figure}

\qift is a quantizer-level replacement for the W2 reconstruction level set in a fixed Hadamard-rotated W2A4/KV4 pipeline. The base pipeline follows the rotation-based PTQ setting~\citep{quarot}: Hadamard transformations redistribute concentrated channel-wise outlier energy before low-bit quantization, which balances the activation distribution and yields the rotated weight source studied in this work (Figure~\ref{fig:activation-max}). This is complementary to activation smoothing such as SmoothQuant~\citep{smoothquant}, which instead migrates activation outlier difficulty into weights.

After rotation, weights may be quantized by RTN-style nearest-level rounding or by compensation methods. RTN selects a per-output-channel scale and rounds each weight to the nearest reconstruction level. GPTQ~\citep{gptq} additionally compensates weight quantization error using a Hessian approximation from calibration activations, and GPTAQ~\citep{gptaq} adds activation-aware asymmetric calibration by matching quantized layer outputs to their full-precision counterparts. These methods reduce the error induced by a chosen four-level codebook, but they do not by themselves determine which four reconstruction levels should be used.

In \qift, the fixed W2 reconstruction level set is isolated as the design intervention. It keeps the standard per-output-channel scale and introduces no group-wise grid assignment, learned codebooks, or asymmetric zero-point metadata. The main \nz\ level set is the uniform four-level mid-rise grid introduced earlier, while \potwide\ keeps the same mirror no-zero structure with power-of-two magnitudes for sign-and-shift decoded weight application. The same level-set replacement can be used with RTN, GPTQ, or GPTAQ, keeping compensation and reconstruction-level geometry as separate design choices. Both level sets are motivated by the source model developed next: a centered, symmetric, Gaussian-like rotated weight distribution should use two inner centroids around zero and two outer centroids in the tails, rather than spending a centroid exactly at zero.

\subsection{Design Principle: A Zero-Centered Gaussian-Like Source}
\label{sec:design-principle}
Reducing weight precision from 16 bits to 2 bits can dramatically reduce memory footprint. However, W2 quantization is qualitatively different from W4 quantization: with only four reconstruction levels, every centroid placement decision matters. In W4, a zero centroid can coexist with many other centroids; in W2, using one centroid at zero consumes 25\% of the representational capacity.

For low-bit quantization, the source distribution strongly affects reconstruction error. If the source is skewed, a sign-symmetric level set wastes levels on the low-density side; if the source is heavy-tailed, the scale or clipping range must cover rare extremes, reducing precision for the dense central region. A zero-centered Gaussian-like source is a useful design reference because it is symmetric, concentrated near the center, and has limited tail dominance. However, exact Gaussianity is not required. The important properties are centeredness, low skewness, moderate kurtosis, and low quantile mismatch to a Gaussian reference.

Concretely, the source hypothesis has two parts:
\begin{align}
    \text{pretrained weights} &\approx \text{near-zero centered}, \\
    \text{Hadamard mixing} &\Rightarrow \text{more Gaussian-like standardized shape}.
\end{align}
If both statements hold, post-rotation weights can be modeled as an approximately zero-centered Gaussian-like source for scalar level-set design. We adopt this model as the design assumption in this section and verify it empirically on real rotated weights in \S\ref{sec:source-validation}, so that the method can be understood independently of the supporting diagnostics.

For a zero-centered Gaussian source, the Lloyd-Max scalar quantization principle~\citep{lloyd,max} implies that the optimal four-level scalar quantizer places two inner centroids around zero and two outer centroids in the tails. It does not allocate a centroid exactly at zero. Since the measured rotated weights are not exactly Gaussian, we use this Lloyd-Max solution as a design prior and scalar reference rather than as a strict generative model.

\begin{figure}[t]
\centering
\figmaybe{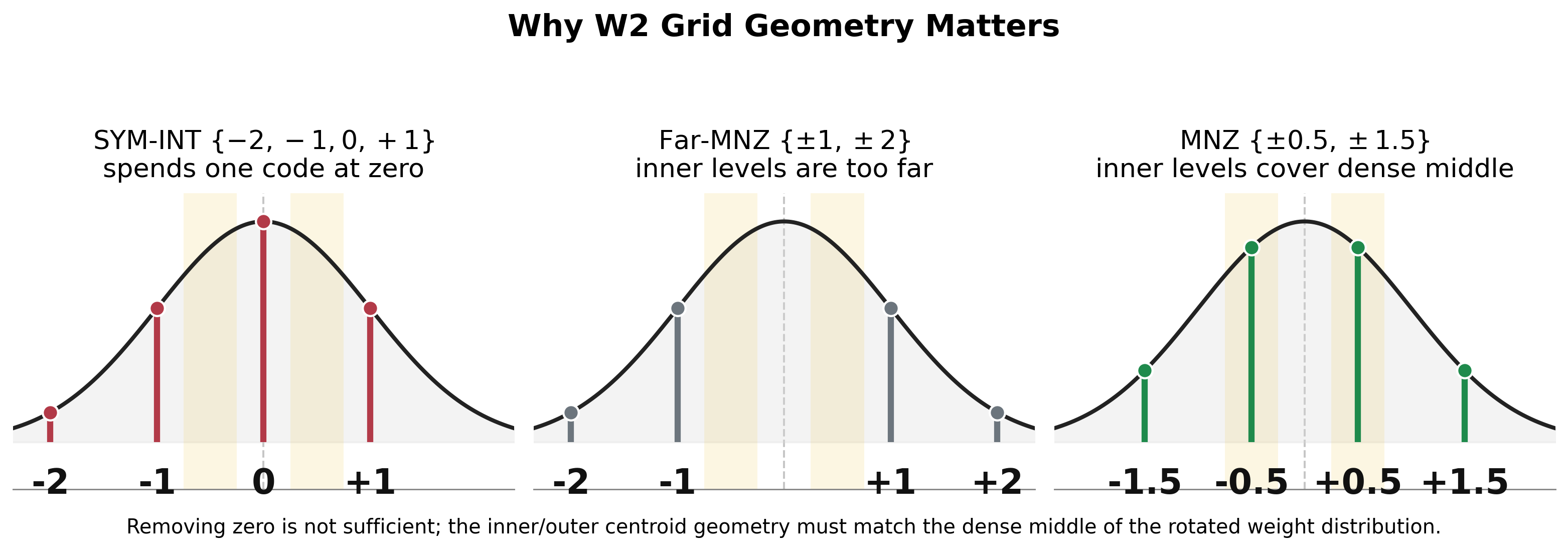}{0.98\linewidth}{Core W2 grid concept.}
\caption{Core intuition behind the proposed W2 grid redesign. The figure illustrates the source model studied in this work: after Hadamard rotation, weights remain approximately zero-centered and become more Gaussian-like in standardized shape. The standard SYM-INT grid spends one reconstruction level exactly at zero, which only benefits very-near-zero weights. \farnz removes the zero level but places the inner centroids too far from zero. \nz places two inner centroids around the dense middle region, yielding a better centroid allocation for the rotated source.}
\label{fig:core-grid-concept}
\end{figure}

The preceding design space suggests that a W2 grid for rotated W2A4/KV4 inference should satisfy four requirements.
First, it should match the post-rotation weight source rather than inherit the standard symmetric integer grid by default.
Second, it should keep the quantizer scalar and globally fixed, avoiding learned partitions, learned codebooks, and group-wise grid assignment.
Third, it should retain the standard per-output-channel scale without introducing asymmetric zero-point metadata.
Fourth, it should remain compatible with simple decoded integer weight application, so that the grid remains accurate and deployment-friendly.
These requirements motivate the concrete no-zero level sets introduced next, whose source assumption is validated empirically in \S\ref{sec:source-validation}.

\subsection{\qift Reconstruction Levels: \nz and \potwide}
The proposed level sets share a common geometry: every level has an equal and
opposite mirror level and no level sits at zero. We call this family
mirror no-zero (MNZ). Figure~\ref{fig:core-grid-concept} illustrates the
intuition: the standard \symint\ grid spends one reconstruction level at zero,
\farnz removes the zero level but places its inner centroids too far from the
dense bulk, and \nz puts its two inner centroids in the dense center of the
rotated distribution. We group the grids into four roles: standard
baselines, proposed \qift\ grids, scalar reference grids, and a negative
diagnostic grid. The proposed deployable grids are \qiftnz and \qiftpot.
Lloyd-Max and NF2 are used as scalar reconstruction references, while \farnz is
included to show that removing zero is not sufficient when the inner centroids
are placed too far from zero. Throughout, we use \nz and \potwide to denote the
level sets themselves, and \qiftnz and \qiftpot to denote the full \qift\ method
instantiated with the corresponding level set.

We evaluate the following W2 grids:
\begin{align}
    \symint &: \{-2,-1,0,+1\}, \\
    \nz &: \{-1.5,-0.5,+0.5,+1.5\}, \\
    \lloyd &: \{\pm0.4528,\pm1.5104\}, \\
    \nf &: \{-1.0,-0.2525685,+0.2525685,+1.0\}, \\
    \potwide &: \{-4,-1,+1,+4\}, \\
    \farnz &: \{-2,-1,+1,+2\}, \quad \text{equivalently } \{\pm1,\pm2\}\ \text{up to scale}\ (r=0.5).
\end{align}

The proposed \nz grid is a drop-in replacement for the reconstruction levels associated with four two-bit codes:
\begin{equation}
    \mathcal{G}_{\mathrm{MNZ}} = \{-1.5,-0.5,+0.5,+1.5\}.
\end{equation}
The grid itself is globally fixed. The same four scalar levels are shared across all layers, modules, channels, and weight groups. Each output channel still uses the standard per-channel scale, but the grid is not learned, tuned, searched, or selected per group. Therefore, \nz introduces no group-wise codebook, no group-wise grid assignment, no per-layer grid search, and no zero-point metadata; the only per-channel quantity is the ordinary scale already used by standard weight quantization.

The same grid can be written as
\begin{equation}
    \{-1.5,-0.5,+0.5,+1.5\}\cdot s
    = \{-3,-1,+1,+3\}\cdot \frac{s}{2}.
\end{equation}
This representation is useful for implementation because the decoded W2 value is an odd integer level, while the half-scale can be folded into the ordinary channel scale. The assignment from two-bit codes to these levels can be chosen by the implementation; the proposed method specifies the reconstruction level set rather than a unique code ordering.

Conventional asymmetric \(b\)-bit quantization represents a weight by an unsigned integer code
\begin{equation}
    q = \mathrm{clip}\!\left(
        \left\lfloor \frac{w}{s} \right\rceil + z,
        0,
        2^b-1
    \right),
    \qquad
    \hat{w}=s(q-z),
\end{equation}
where \(z\) is a zero-point. For \(b=2\), the unsigned code set is \(q\in\{0,1,2,3\}\). For nearly symmetric rotated weights, the useful midpoint between the four integer codes is \(z=1.5\). Keeping this fractional midpoint gives
\begin{equation}
    s\{0-1.5,1-1.5,2-1.5,3-1.5\}
    = s\{-1.5,-0.5,+0.5,+1.5\},
\end{equation}
which is exactly \nz. Thus, \nz can be viewed as a zero-point-free realization of the useful fractional-midpoint geometry of asymmetric W2. It preserves the favorable no-zero level placement without storing per-channel zero-points or performing zero-point subtraction during dequantization.

To further reduce arithmetic cost, we also study the power-of-two no-zero grid
\begin{equation}
    \mathcal{G}_{\mathrm{pot}}=\{\pm1,\pm4\}.
\end{equation}

\subsection{Inner/Outer Ratio as a Design Knob}
Beyond removing the zero centroid, the main scale-invariant degree of freedom is the inner/outer centroid ratio. Normalizing the outer magnitude to one, the mirror no-zero family is
\begin{equation}
\mathcal{G}(r)=\{-1,-r,+r,+1\}, \quad 0<r<1.
\end{equation}
The grids studied here correspond to different ratios: \qiftnz to $r=1/3$, \qiftpot to $r=1/4$, the Lloyd-Max and NF2 references to $r\approx0.30$ and $r\approx0.25$, and the negative diagnostic \farnz to $r=1/2$. A scale-invariant reconstruction analysis on real rotated weights (\S\ref{sec:ratio-sensitivity}) shows that effective grids cluster in the band $r\approx0.25$--$0.33$, which contains \nz, \potwide, Lloyd-Max, and NF2 but excludes \farnz. We therefore treat $r$ as the primary geometric design knob and choose grids inside this band.

\subsection{Quantization Objective and Integration with GPTQ/GPTAQ}
Since \qift is a post-training grid replacement, it introduces no trainable objective. The only local objective is the standard per-output-channel scale selection that minimizes nearest-level reconstruction error for the chosen grid, identical to the scale search used by the baseline quantizer. When GPTQ or GPTAQ is enabled, \qift uses the same Hessian-based or asymmetric-calibration compensation as the baseline pipeline; only the fixed reconstruction level set changes, and no additional parameters, learned centroids, or zero-points are introduced.

Operationally, the standard W2 grid is replaced by a fixed no-zero grid while the surrounding rotated PTQ pipeline is unchanged. For each output channel, the quantizer selects an MSE-optimal scale for the chosen grid, maps weights to the nearest scaled level, and applies the same GPTQ or GPTAQ compensation as the baseline pipeline.

\subsection{Hardware-Friendly Decoded Levels}
The proposed level sets also have a simple decoded arithmetic structure. \qiftnz can be represented as odd integer levels $\{\pm1,\pm3\}$ with a half-scale reparameterization, while \qiftpot uses power-of-two levels $\{\pm1,\pm4\}$. Thus, after decoding, the weight values are small fixed signed integers rather than learned or irregular lookup-table values. This regular integer structure makes the level sets hardware-friendly.

\section{Experiments}

\subsection{Experimental Setup}

We evaluate LLaMA-2-7B and LLaMA-3.1-8B. The main stress test is pure W2A4, where all linear layers use two-bit weights and four-bit activations. We also evaluate the L-layer W2/W4 mixed-precision heuristic described in the mixed-L section, especially the $L=16$ iso-bit point where the nominal average weight bit-width is three bits. We report WikiText-2 perplexity and downstream accuracy over ARC-Challenge, ARC-Easy, HellaSwag, PIQA, and WinoGrande.

\paragraph{Quantization configuration.}
Unless otherwise stated, all experiments use a QuaRot-style Hadamard-rotated pipeline with weight, activation, and KV-cache quantization enabled. For the W2A4/KV4 setting, weights are quantized to two bits while activations and the KV-cache use four bits. Weights use per-output-channel scaling without grouping, and the per-channel scale is selected by a clipping-based search. The conventional asymmetric baseline additionally enables asymmetric weight quantization with a stored per-channel zero-point, whereas \qift\ level sets change only the fixed two-bit code-to-level mapping and remain zero-point-free. Activations use symmetric four-bit quantization with a clipping ratio of $0.9$, and the KV-cache uses asymmetric key and value quantization with clipping ratios of $0.95$. We use the default Hadamard rotation and do not learn or sample random rotation matrices. GPTQ uses a small Hessian damping factor ($0.01$), and GPTAQ uses the same quantization configuration and calibration set with activation-aware asymmetric correction additionally enabled.

\paragraph{Calibration, evaluation, and software.}
We use 128 WikiText-2 calibration samples with sequence length 2048 and sampling seed 0.
Perplexity is evaluated on WikiText-2, and downstream accuracy is evaluated on ARC-Challenge, ARC-Easy, HellaSwag, PIQA, and WinoGrande using the same task suite across grid variants.
All reported results are single-seed unless otherwise specified.
FP16 reference rows use W16A16/KV16 without weight, activation, or KV-cache quantization; they provide the full-precision upper-bound reference for the same WikiText-2 and downstream task suite.
Experiments are run in the project Docker environment with Python 3.10.13, PyTorch 2.2.1, CUDA 12.1, and NVIDIA RTX A6000 48GB GPUs.

\subsection{Source-Model Validation}
\label{sec:source-validation}
Before evaluating the proposed grids, we empirically verify the source assumption adopted in \S\ref{sec:design-principle}: that Hadamard-rotated weights are approximately zero-centered and become more Gaussian-like in standardized shape. We use the term Gaussian-like operationally---a distribution is more Gaussian-like if its standardized shape has lower absolute skewness, lower absolute excess kurtosis, and lower Q--Q error against a standard normal reference---rather than as a claim that the weights exactly follow a normal law.

For each output channel $c$ with weights $W_c$, define
\begin{equation}
    r_c = \frac{|\mu_c|}{\sigma_c}, \qquad
    \mu_c = \mathbb{E}[W_c], \quad \sigma_c = \sqrt{\mathrm{Var}(W_c)}.
\end{equation}
This measures centeredness. To measure shape, we standardize weights as $z=(w-\mu)/\sigma$ and compute:
\begin{align}
    \text{excess kurtosis} &= \mathbb{E}[z^4]-3, \\
    \text{skewness} &= \mathbb{E}[z^3], \\
    \text{Q--Q error} &= \frac{1}{m}\sum_{i=1}^{m}\left(q_i^{\mathrm{emp}} - q_i^{\mathcal{N}(0,1)}\right)^2.
\end{align}
For a standard Gaussian, excess kurtosis and skewness are zero, and the Q--Q error is low. We use these metrics as diagnostics of Gaussian-likeness, not as a claim that the empirical distribution exactly follows a normal law.

\begin{figure}[t]
\centering
\figmaybe{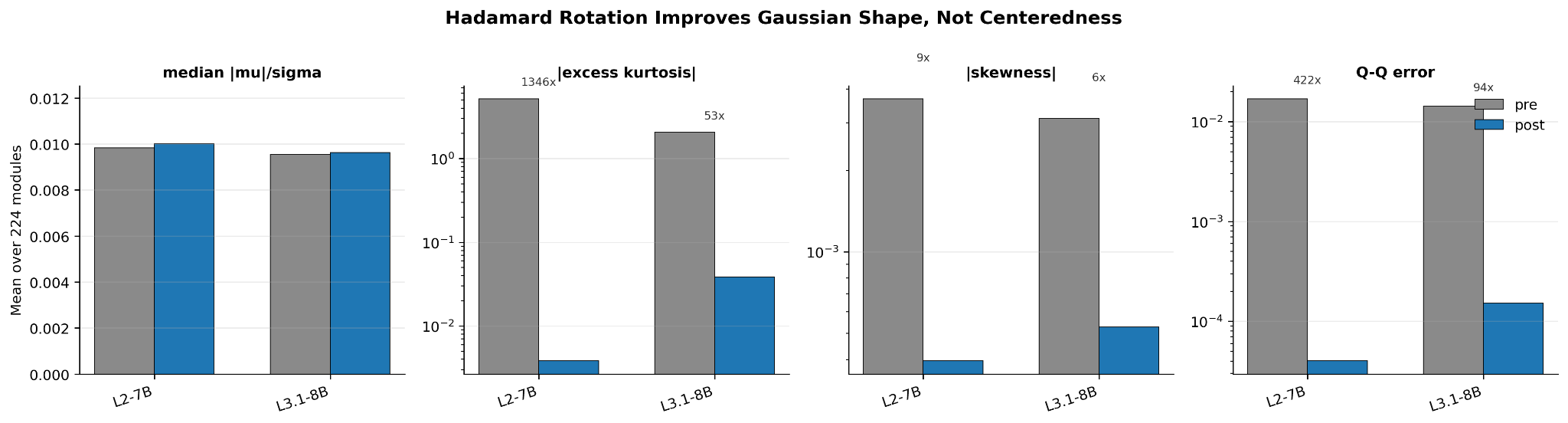}{0.95\linewidth}{Gaussianity summary.}
\caption{Hadamard rotation improves kurtosis and Q--Q shape metrics by orders of magnitude (skewness more modestly) while leaving mean-centeredness essentially unchanged.}
\label{fig:gaussian-summary}
\end{figure}

Figure~\ref{fig:gaussian-summary} and Table~\ref{tab:gaussian-summary} show model-wide pre/post statistics across 224 linear modules. The centeredness metric remains nearly unchanged, confirming that Hadamard rotation does not create zero-centeredness. In contrast, excess kurtosis, skewness, and Q--Q error drop sharply, showing that rotation primarily makes standardized shape more Gaussian-like.

\begin{table}[t]
\centering
\caption{Model-wide Gaussianity summary across 224 linear modules. Lower is better for all metrics shown.}
\label{tab:gaussian-summary}
\begin{tabular}{llccc}
\toprule
Model & Metric & Pre & Post & Change \\
\midrule
LLaMA-2-7B & mean $|\text{excess kurtosis}|$ & 5.164 & 0.00384 & $-99.93\%$ \\
LLaMA-2-7B & mean Q--Q error & 0.01693 & 0.0000401 & $-99.76\%$ \\
LLaMA-2-7B & mean $|\text{skewness}|$ & 0.00368 & 0.000395 & $-89.25\%$ \\
LLaMA-2-7B & mean median $|\mu|/\sigma$ & 0.00982 & 0.01002 & unchanged \\
\midrule
LLaMA-3.1-8B & mean $|\text{excess kurtosis}|$ & 2.057 & 0.0387 & $-98.12\%$ \\
LLaMA-3.1-8B & mean Q--Q error & 0.01429 & 0.000152 & $-98.93\%$ \\
LLaMA-3.1-8B & mean $|\text{skewness}|$ & 0.00311 & 0.000527 & $-83.07\%$ \\
LLaMA-3.1-8B & mean median $|\mu|/\sigma$ & 0.00953 & 0.00963 & unchanged \\
\bottomrule
\end{tabular}
\end{table}

The improvement is not caused by a small number of selected modules: across modules the post-rotation shape metrics improve for the large majority, and the improvement is consistent across all 32 layers.

\subsection{Main Performance Results}
We first establish the failure of the standard grid under pure W2A4 and then show that source-aware no-zero grids recover most of the lost accuracy. Table~\ref{tab:summary} summarizes the headline comparison across both models and the three main operating points; the remainder of this section reports the full per-grid and per-task breakdowns that support it.

\begin{table}[t]
\centering
\caption{Headline summary of \qift\ versus baselines under rotated KV4 inference. ``Down.\ Avg'' is the mean zero-shot accuracy over ARC-C, ARC-E, HellaSwag, PIQA, and WinoGrande. \emph{Pure W2A4} keeps all linear layers at two-bit weights; \emph{$L{=}16$} upgrades the 16 most sensitive layers to W4A4, giving \(b_{\mathrm{avg}}=3\), iso-bit with uniform W3A4. Lower PPL and higher accuracy are better. Detailed per-grid and per-task results appear in the tables that follow.}
\label{tab:summary}
\small
\setlength{\tabcolsep}{4pt}
\begin{tabularx}{\linewidth}{l l l r r >{\raggedright\arraybackslash}X}
\toprule
Model & Setting & Method & PPL$\downarrow$ & Down.\ Avg$\uparrow$ & Notes \\
\midrule
\multirow{5}{*}{LLaMA-2-7B}
 & FP16              & --       & 5.471  & 0.6866 & upper bound \\
 & Pure W2A4 (GPTAQ) & \symint  & 12.118 & 0.4211 & GPTQ collapses (53.849 PPL) \\
 & Pure W2A4 (GPTAQ) & \qiftnz  & 9.294  & 0.4794 & fixed no-zero recovers \\
\cmidrule(lr){2-6}
 & $L{=}16$ (GPTAQ)  & \qiftnz  & 7.122  & 0.6157 & iso-bit with W3A4 \\
 & W3A4 (GPTQ)       & \symint  & 6.897  & 0.6200 & iso-bit reference \\
\midrule
\multirow{5}{*}{LLaMA-3.1-8B}
 & FP16              & --       & 6.277  & 0.7435 & upper bound \\
 & Pure W2A4 (GPTAQ) & \symint  & 29.695 & 0.3683 & GPTQ collapses ($>$3000 PPL) \\
 & Pure W2A4 (GPTAQ) & \qiftnz  & 19.515 & 0.4064 & fixed no-zero recovers \\
\cmidrule(lr){2-6}
 & $L{=}16$ (GPTAQ)  & \qiftnz  & 11.936 & 0.5609 & narrows W3A4 gap \\
 & W3A4 (GPTQ)       & \symint  & 10.954 & 0.5972 & iso-bit reference \\
\bottomrule
\end{tabularx}
\end{table}

Under pure quantization, where all linear layers are quantized to W2A4 with KV4 and GPTQ, the standard \symint\ W2 grid is severe: LLaMA-2-7B degrades from an FP16 reference of 5.471 to 53.849 PPL, while LLaMA-3.1-8B collapses completely (FP16 6.277 $\rightarrow$ 3005.556 PPL).

\begin{figure}[t]
\centering
\figmaybe{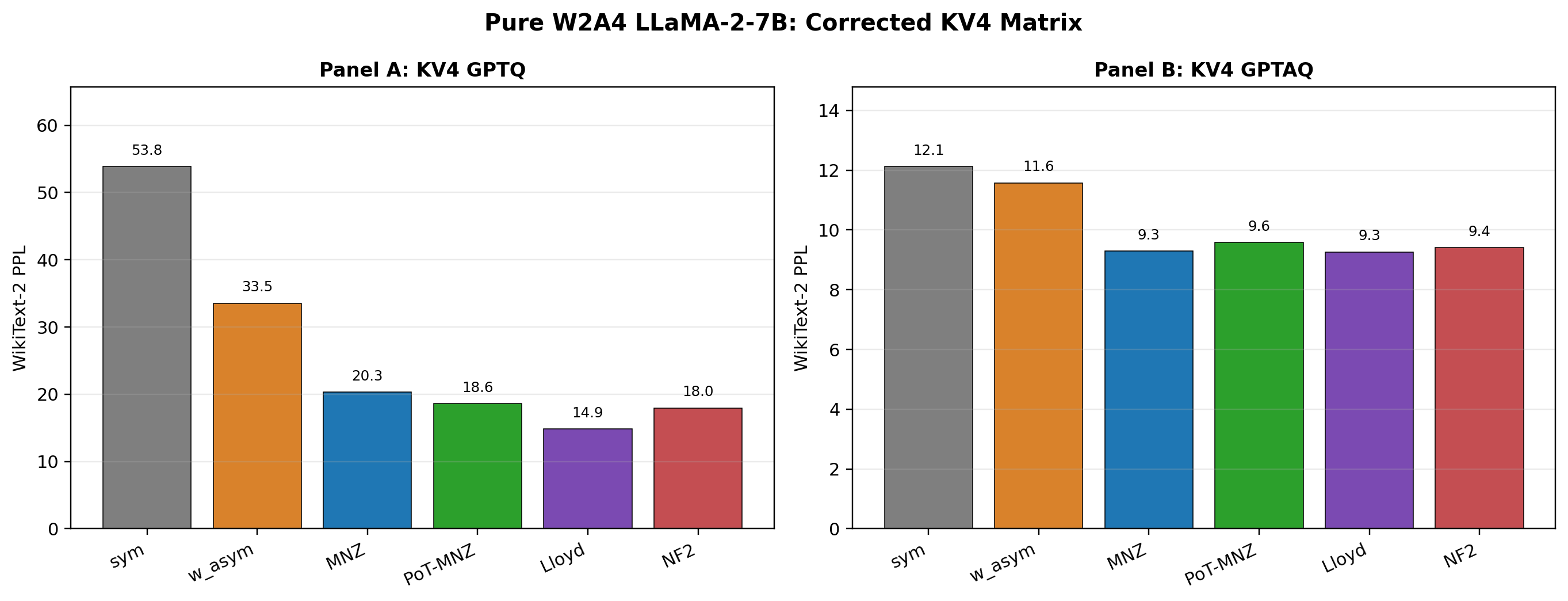}{0.95\linewidth}{LLaMA-2 pure W2A4 bar chart.}
\caption{Pure W2A4 KV4 results on LLaMA-2-7B. No-zero grids substantially outperform standard W2 and \wasym.}
\label{fig:l2-pure-bar}
\end{figure}

\begin{figure}[t]
\centering
\figmaybe{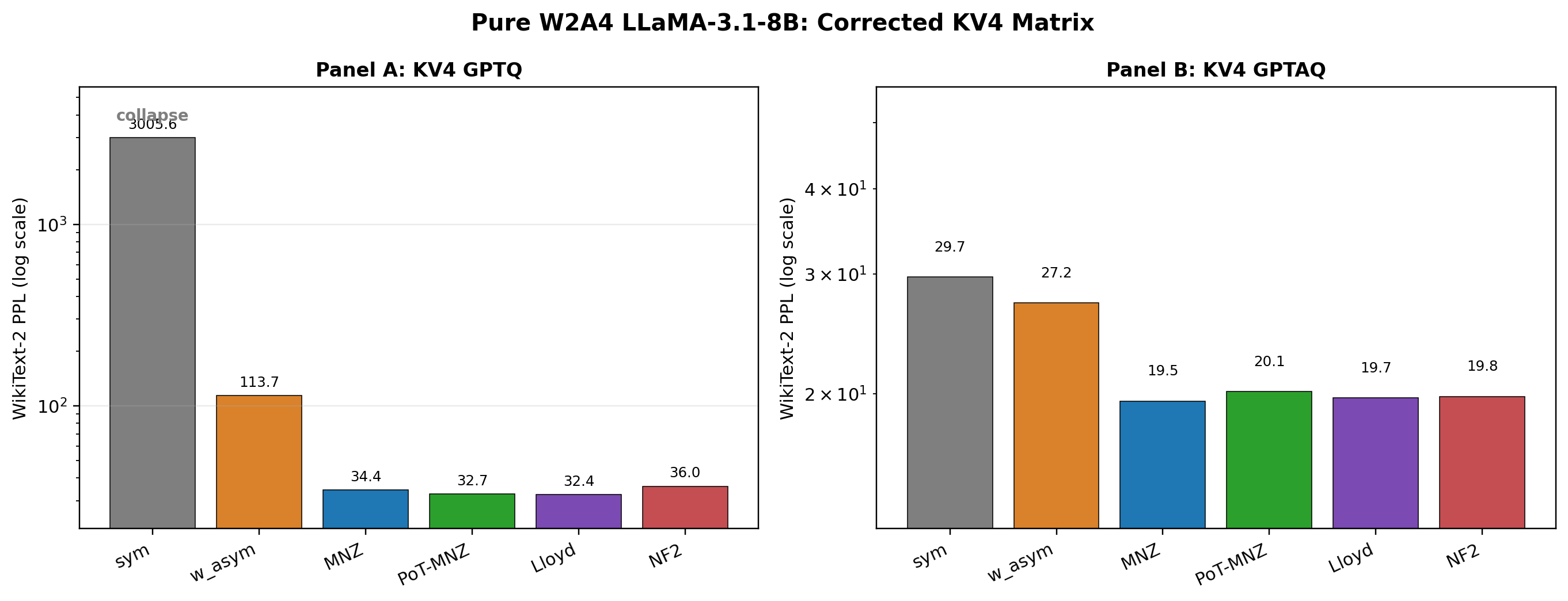}{0.95\linewidth}{LLaMA-3.1 pure W2A4 bar chart.}
\caption{Pure W2A4 KV4 results on LLaMA-3.1-8B. The standard \symint\ grid collapses, while no-zero grids recover the model to a usable range.}
\label{fig:l31-pure-bar}
\end{figure}

Tables~\ref{tab:l2-pure} and~\ref{tab:l31-pure} summarize pure W2A4 results. Figures~\ref{fig:l2-pure-bar} and~\ref{fig:l31-pure-bar} visualize the KV4 matrix. We read these results in three steps: asymmetric W2 improves over the standard \symint\ baseline, confirming that scalar level-set geometry matters; Gaussian-aware no-zero grids further improve over asymmetric W2, showing that source-aware centroid placement is more effective than generic zero-point correction; and \potwide remains competitive while using power-of-two reconstruction levels that map naturally to sign-and-shift weight application. Lloyd-Max is often the strongest scalar reference, as expected from the Gaussian source model, while \qiftnz\ and \qiftpot\ remain close with simpler integer or power-of-two decoded levels. These improvements are obtained without changing the calibration procedure, learning a codebook, introducing per-group grids, or storing zero-points; the only grid-level change is the fixed W2 code-to-level mapping.

\begin{table}[t]
\centering
\caption{Pure W2A4 LLaMA-2-7B corrected matrix. Lower PPL is better. \qiftnz and \qiftpot are the proposed grids; Lloyd-Max and NF2 are scalar reference grids. Bold marks the best non-reference result in each column.}
\label{tab:l2-pure}
\begin{tabular}{lrrrr}
\toprule
Grid & KV4 GPTQ & KV4 GPTAQ & KV16 GPTQ & KV16 GPTAQ \\
\midrule
FP16 & \multicolumn{4}{c}{5.471} \\
sym & 53.849 & 12.118 & 51.242 & 12.007 \\
w\_asym & 33.533 & 11.577 & 31.271 & 10.848 \\
\qiftnz & 20.297 & 9.294 & 18.690 & 9.078 \\
\qiftpot & 18.580 & 9.577 & 17.208 & 9.275 \\
Lloyd-Max & \textbf{14.856} & \textbf{9.265} & \textbf{13.983} & \textbf{8.921} \\
NF2 & 17.966 & 9.404 & 16.826 & 9.084 \\
\bottomrule
\end{tabular}
\end{table}

\begin{table}[t]
\centering
\caption{Pure W2A4 LLaMA-3.1-8B corrected matrix. Lower PPL is better. \qiftnz and \qiftpot are the proposed grids; Lloyd-Max and NF2 are scalar reference grids.}
\label{tab:l31-pure}
\begin{tabular}{lrrrr}
\toprule
Grid & KV4 GPTQ & KV4 GPTAQ & KV16 GPTQ & KV16 GPTAQ \\
\midrule
FP16 & \multicolumn{4}{c}{6.277} \\
sym & 3005.556 & 29.695 & 3593.113 & 26.573 \\
w\_asym & 113.747 & 27.197 & 93.398 & 25.342 \\
\qiftnz & 34.396 & \textbf{19.515} & 27.794 & 17.810 \\
\qiftpot & 32.693 & 20.150 & 23.783 & 17.281 \\
Lloyd-Max & \textbf{32.362} & 19.724 & \textbf{23.222} & \textbf{17.261} \\
NF2 & 36.034 & 19.832 & 26.058 & 17.307 \\
\bottomrule
\end{tabular}
\end{table}

Table~\ref{tab:pure-downstream} shows pure W2A4 downstream results on LLaMA-2-7B under KV4. No-zero grids improve substantially over both standard \symint\ W2 and asymmetric W2.

\begin{table}[t]
\centering
\caption{Pure W2A4 KV4 downstream accuracy on LLaMA-2-7B. \qiftnz and \qiftpot are the proposed grids; Lloyd-Max and NF2 are scalar reference grids. Higher is better. Bold marks the best non-reference result.}
\label{tab:pure-downstream}
\begin{tabular}{lcccccc}
\toprule
Setting & ARC-C & ARC-E & HellaSwag & PIQA & WinoGrande & Avg \\
\midrule
FP16 & 0.4514 & 0.7382 & 0.7619 & 0.7878 & 0.6938 & 0.6866 \\
W4A4 GPTQ & 0.4164 & 0.6835 & 0.7234 & 0.7639 & 0.6448 & 0.6464 \\
W2A4 sym GPTQ & 0.2253 & 0.3182 & 0.3030 & 0.5348 & 0.4941 & 0.3751 \\
W2A4 sym GPTAQ & 0.2338 & 0.3826 & 0.3773 & 0.5903 & 0.5217 & 0.4211 \\
W2A4 w\_asym GPTAQ & 0.2355 & 0.3859 & 0.3876 & 0.5914 & 0.5296 & 0.4260 \\
W2A4 \qiftnz{} GPTAQ & 0.2816 & 0.4659 & 0.4648 & 0.6425 & 0.5422 & \textbf{0.4794} \\
W2A4 \qiftpot{} GPTAQ & 0.2756 & 0.4625 & 0.4453 & 0.6355 & 0.5643 & 0.4766 \\
W2A4 Lloyd-Max GPTAQ & 0.2611 & 0.4722 & 0.4674 & 0.6235 & 0.5722 & 0.4793 \\
W2A4 NF2 GPTAQ & 0.2773 & 0.4659 & 0.4567 & 0.6491 & 0.5588 & 0.4816 \\
\bottomrule
\end{tabular}
\end{table}

Table~\ref{tab:l31-downstream} reports the same downstream task suite on LLaMA-3.1-8B under KV4. The trend from LLaMA-2 transfers to the newer model: Gaussian-aware W2 grids improve both pure W2A4 GPTAQ and L=16 GPTAQ. \potwide is the best average point in both LLaMA-3.1 W2 groups, supporting the ratio-sensitivity argument that its $r=0.25$ geometry is not merely hardware-convenient.

\begin{table}[t]
\centering
\caption{LLaMA-3.1-8B KV4 downstream accuracy on the evaluation task subset. \qiftnz and \qiftpot are the proposed grids; Lloyd-Max is a scalar reference grid. Higher is better. Bold marks the best non-reference result within each comparison block.}
\label{tab:l31-downstream}
\begin{tabular}{lcccccc}
\toprule
Setting & ARC-C & ARC-E & HellaSwag & PIQA & WinoGrande & Avg \\
\midrule
FP16 & 0.5495 & 0.8220 & 0.7941 & 0.8069 & 0.7451 & 0.7435 \\
W4A4 GPTQ & 0.4317 & 0.6810 & 0.7242 & 0.7470 & 0.6638 & 0.6495 \\
W3A4 GPTQ & 0.3797 & 0.6296 & 0.6529 & 0.7111 & 0.6125 & 0.5972 \\
\midrule
W2A4 sym GPTAQ & 0.2167 & 0.2980 & 0.3021 & 0.5163 & 0.5083 & 0.3683 \\
W2A4 \qiftnz{} GPTAQ & 0.2329 & 0.3556 & 0.3650 & 0.5598 & 0.5185 & 0.4064 \\
W2A4 \qiftpot{} GPTAQ & 0.2329 & 0.3712 & 0.3755 & 0.5609 & 0.5296 & \textbf{0.4140} \\
W2A4 Lloyd-Max GPTAQ & 0.2201 & 0.3569 & 0.3730 & 0.5511 & 0.5335 & 0.4069 \\
\midrule
L16 sym GPTAQ & 0.3114 & 0.5417 & 0.5794 & 0.6632 & 0.6196 & 0.5431 \\
L16 \qiftnz{} GPTAQ & 0.3302 & 0.5724 & 0.6047 & 0.6904 & 0.6069 & 0.5609 \\
L16 \qiftpot{} GPTAQ & 0.3439 & 0.5859 & 0.6043 & 0.6785 & 0.5967 & \textbf{0.5619} \\
L16 Lloyd-Max GPTAQ & 0.3387 & 0.5657 & 0.6081 & 0.6861 & 0.5959 & 0.5589 \\
\bottomrule
\end{tabular}
\end{table}

On pure W2A4, \potwide improves LLaMA-3.1 average accuracy from 0.3683 to 0.4140. In L=16 mixed precision, \potwide improves from 0.5431 to 0.5619 and remains close to the W3A4 reference at 0.5972. These results strengthen the cross-model claim: the grid redesign is not specific to LLaMA-2, and the hardware-friendly \potwide variant remains competitive under downstream metrics.

\subsection{Constraint-Specific Comparison: $L{=}16$ vs.\ W3A4}

Before reporting mixed-L results, we describe the simple layer-selection heuristic used in this experiment. The mixed-L setting is a deployment-oriented stress test for the proposed W2 level-set redesign, not a general mixed-precision allocation algorithm. The goal is to ask whether W2A4 can retain much of its storage advantage while protecting a small number of highly sensitive layers.

Starting from a pure W2A4 baseline, we independently upgrade one transformer layer at a time to W4A4 while keeping all other layers at W2A4. Let $P_0$ denote the perplexity of the pure W2A4 baseline and $P_\ell$ denote the perplexity after upgrading only layer $\ell$ to W4A4. We define the single-layer gain as
\begin{equation}
    \Delta_\ell = P_0 - P_\ell .
\end{equation}
Layers are sorted by $\Delta_\ell$, and the top-$L$ layers are upgraded to W4A4 in the final mixed-precision model; all remaining layers stay at W2A4.

For a 32-layer model, upgrading one full layer from W2 to W4 increases the average weight bit-width by approximately $2/32$. Thus the nominal average weight bit-width is
\begin{equation}
    b_{\mathrm{avg}} = 2 + \frac{2L}{32} = 2 + \frac{L}{16}.
\end{equation}
At $L=16$, this gives $b_{\mathrm{avg}}=3$, matching the nominal average weight budget of uniform W3A4.

This layer-wise mixed-precision setting also keeps the quantization layout regular. Sub-tensor mixed precision can protect outlier channels or elements, but then a single matrix multiplication may contain multiple weight precisions. Our policy is deliberately layer-wise: selected sensitive layers are assigned W4A4, while the remaining layers stay at uniform W2A4. Thus every quantized matrix multiplication keeps a single weight precision.

This heuristic is intentionally simple. It ignores joint interactions between upgraded layers and is not claimed to be optimal. We use it only to construct a transparent W2/W4 Pareto curve and to evaluate whether the proposed no-zero W2 grids remain useful when only part of the model stays at two-bit precision.

\begin{figure}[t]
\centering
\figmaybe{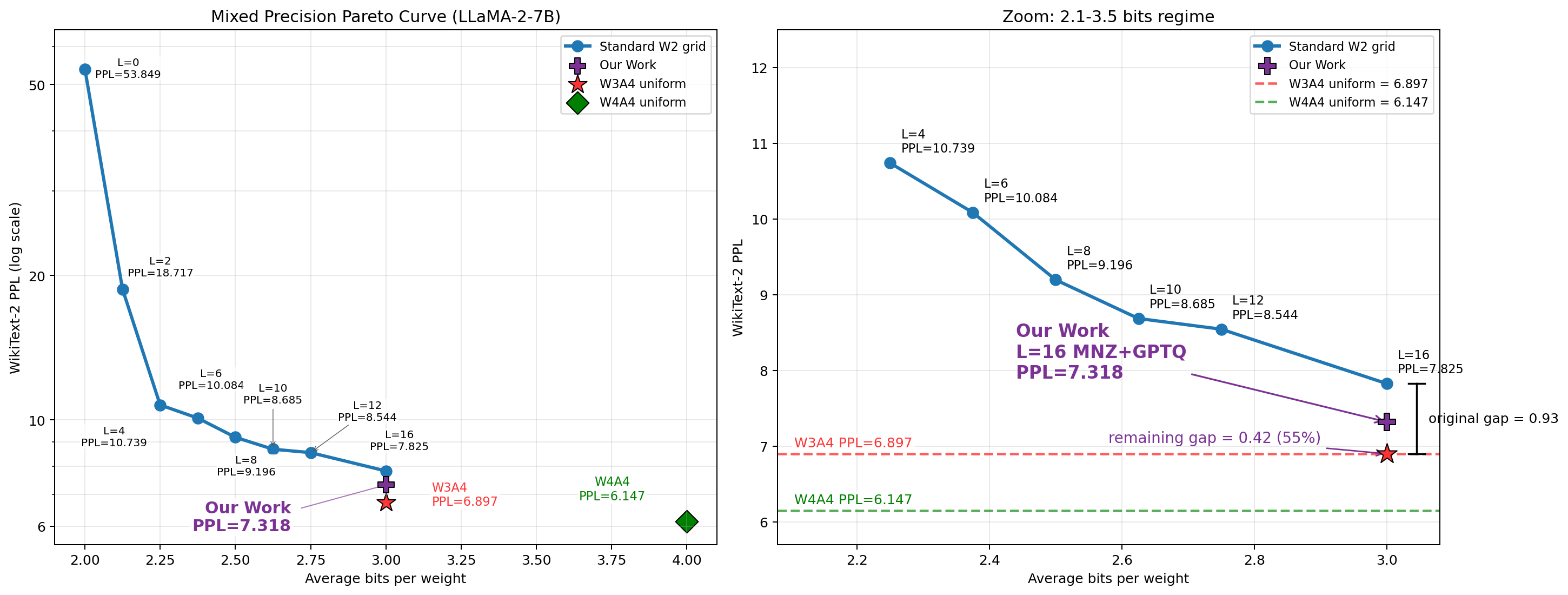}{0.98\linewidth}{L-layer W2/W4 mixed precision Pareto curve.}
\caption{Simple W2/W4 mixed-precision Pareto curve on LLaMA-2-7B. Starting from pure W2A4, layers are ranked by the perplexity gain from independently upgrading each layer to W4A4, and then the top-$L$ layers are upgraded. The average weight bit-width is $2+L/16$. At $L=16$, the model reaches the same nominal average weight bit-width as W3A4. The purple marker shows the iso-method L=16 MNZ point at 7.318 PPL under the same GPTQ calibration as the standard-grid curve. Under iso-method comparison, the level-set redesign alone (\symint$\to$MNZ) closes about 55\% of the W3A4 gap under GPTQ (7.825$\to$7.318 against 6.897) and about 49\% under GPTAQ (7.719$\to$7.122 against 6.509); in both cases L=16 approaches but does not surpass W3A4.}
\label{fig:l-mixed-pareto}
\end{figure}

Figure~\ref{fig:l-mixed-pareto} shows the W2/W4 mixed-precision Pareto curve obtained from the single-layer sensitivity ranking. Most of the recovery occurs in the first few selected layers: L=4 already reduces perplexity from 53.849 to 10.739, and the curve has a knee around L=4--6. This indicates that standard W2A4 failure is concentrated in a small number of functionally sensitive layers.

Tables~\ref{tab:l2-mixed-l-ppl} and~\ref{tab:l31-mixed-l-ppl} report the corresponding mixed-L WikiText-2 perplexity matrix under KV4 GPTQ. After rerunning corrected NF2 for L=4/8/12, the LLaMA-2 table is complete for all six grids: standard \symint\ W2, conventional asymmetric W2, \nz, \potwide, Lloyd, and corrected NF2. The LLaMA-3.1 table is also complete for all six grids. For LLaMA-3.1, we intentionally reuse the same static L-layer ranking from the LLaMA-2 sweep rather than performing a model-specific L-search; this treats mixed-L selection as a fixed deployment heuristic and tests whether the W2 grid trend transfers across models.

\begin{table}[t]
\centering
\caption{LLaMA-2-7B mixed-L KV4 GPTQ WikiText-2 perplexity. \qiftnz and \qiftpot are the proposed grids; Lloyd-Max and NF2 are scalar reference grids. Lower is better.}
\label{tab:l2-mixed-l-ppl}
\begin{tabular}{lrrrrrrr}
\toprule
L & Avg bits & sym & w\_asym & \qiftnz & \qiftpot & Lloyd-Max & NF2 \\
\midrule
FP16 & 16.000 & \multicolumn{6}{c}{5.471} \\
W3A4 ref. & 3.000 & \multicolumn{6}{c}{6.897} \\
4  & 2.250 & 10.739 & 11.189 & 9.273 & 9.810 & \textbf{9.242} & 9.799 \\
8  & 2.500 & 9.196  & 9.129  & \textbf{8.144} & 8.231 & 8.191 & 8.220 \\
12 & 2.750 & 8.544  & 8.468  & \textbf{7.673} & 7.732 & 7.684 & 7.728 \\
16 & 3.000 & 7.825  & 7.831  & 7.318 & 7.316 & \textbf{7.278} & 7.325 \\
\bottomrule
\end{tabular}
\end{table}

\begin{table}[t]
\centering
\caption{LLaMA-3.1-8B mixed-L KV4 GPTQ WikiText-2 perplexity using the same static L-layer ranking as the LLaMA-2 sweep. \qiftnz and \qiftpot are the proposed grids; Lloyd-Max and NF2 are scalar reference grids. Lower is better.}
\label{tab:l31-mixed-l-ppl}
\begin{tabular}{lrrrrrrr}
\toprule
L & Avg bits & sym & w\_asym & \qiftnz & \qiftpot & Lloyd-Max & NF2 \\
\midrule
FP16 & 16.000 & \multicolumn{6}{c}{6.277} \\
W3A4 ref. & 3.000 & \multicolumn{6}{c}{10.954} \\
4  & 2.250 & 21.203 & 20.330 & 17.801 & 17.467 & \textbf{15.814} & 17.061 \\
8  & 2.500 & 17.120 & 17.231 & 14.330 & 14.557 & 14.361 & \textbf{14.326} \\
12 & 2.750 & 14.557 & 14.042 & \textbf{12.443} & 12.665 & 12.603 & 12.468 \\
16 & 3.000 & 12.744 & 12.510 & 11.521 & \textbf{11.499} & 11.537 & 11.548 \\
\bottomrule
\end{tabular}
\end{table}

The mixed-L results show that level-set redesign remains useful beyond pure W2A4. On LLaMA-2, \nz, \potwide, Lloyd, and corrected NF2 form a tight cluster for L=4--16, with Lloyd slightly best at L=4/16 and \nz best at L=8/12. Corrected NF2 is competitive but not the best LLaMA-2 mixed-L grid. On LLaMA-3.1, the same trend transfers: no-zero and Gaussian-aware grids consistently improve over the standard \symint\ grid, and \potwide becomes the best L=16 point despite being the hardware-friendly power-of-two variant.

Table~\ref{tab:mixed-l-gptaq-ppl} further evaluates pure W2A4 and mixed-L under GPTAQ, including the conventional asymmetric W2 baseline. L=0 denotes the pure W2A4 setting under GPTAQ. This should not be confused with the earlier pure W2A4 GPTQ collapse baseline: for example, LLaMA-2-7B pure W2A4 symmetric quantization is 53.849 PPL with GPTQ but 12.118 PPL with GPTAQ in this table. The main conclusion is not that GPTAQ is universally beneficial. Rather, the interaction is model-dependent: on LLaMA-2-7B, GPTAQ lowers the L=16 endpoints for all shown grids relative to the GPTQ table, while on LLaMA-3.1-8B the same GPTAQ sweep worsens the L=16 endpoints. In both models, however, the no-zero/Gaussian-aware grids remain consistently better than the standard \symint\ grid and the conventional asymmetric baseline across all L. This separates the robust level-set-geometry effect from the less stable GPTAQ interaction.

\begin{table}[t]
\centering
\caption{Pure W2A4 and mixed-L KV4 GPTAQ WikiText-2 perplexity on LLaMA-2-7B and LLaMA-3.1-8B. \qiftnz and \qiftpot are the proposed grids; Lloyd-Max is a scalar reference grid. Lower is better. L=0 denotes pure W2A4 with no W4A4 protected layers, evaluated with GPTAQ rather than GPTQ.}
\label{tab:mixed-l-gptaq-ppl}
\small
\setlength{\tabcolsep}{4pt}
\begin{tabular}{llrrrrrr}
\toprule
Model & L & Avg bits & sym & w\_asym & \qiftnz & \qiftpot & Lloyd-Max \\
\midrule
LLaMA-2-7B & FP16 & 16.000 & \multicolumn{5}{c}{5.471} \\
LLaMA-2-7B & W3A4 ref. & 3.000 & \multicolumn{5}{c}{6.509} \\
LLaMA-2-7B & 0  & 2.000 & 12.118 & 11.577 & 9.294 & 9.577 & \textbf{9.265} \\
LLaMA-2-7B & 4  & 2.250 & 10.466 & 9.776 & 8.426 & 8.531 & \textbf{8.422} \\
LLaMA-2-7B & 8  & 2.500 & 9.243  & 9.123 & \textbf{8.009} & 8.124 & 8.015 \\
LLaMA-2-7B & 12 & 2.750 & 8.484  & 8.267 & 7.565 & 7.651 & \textbf{7.552} \\
LLaMA-2-7B & 16 & 3.000 & 7.719  & 7.572 & \textbf{7.122} & 7.164 & 7.145 \\
\midrule
LLaMA-3.1-8B & FP16 & 16.000 & \multicolumn{5}{c}{6.277} \\
LLaMA-3.1-8B & 0  & 2.000 & 29.695 & 27.197 & \textbf{19.515} & 20.150 & 19.724 \\
LLaMA-3.1-8B & 4  & 2.250 & 23.345 & 21.901 & 17.171 & 17.365 & \textbf{16.625} \\
LLaMA-3.1-8B & 8  & 2.500 & 19.085 & 17.976 & 14.911 & 14.841 & \textbf{14.798} \\
LLaMA-3.1-8B & 12 & 2.750 & 15.628 & 15.243 & 13.320 & 13.359 & \textbf{13.222} \\
LLaMA-3.1-8B & 16 & 3.000 & 13.683 & 13.226 & 11.936 & 12.056 & \textbf{11.882} \\
\bottomrule
\end{tabular}
\end{table}

Under WikiText-2 perplexity, L=16 mixed precision approaches but does not fully surpass the uniform W3A4 reference under the strict KV4 iso-bit comparison. To isolate the level-set effect from the calibration-algorithm effect, we compare each grid against the W3A4 reference under the \emph{same} calibration method. Under GPTQ, the standard \symint\ L=16 point is 7.825 PPL versus a W3A4 GPTQ reference of 6.897 PPL, a 0.928 PPL gap; replacing \symint\ with \nz lowers the L=16 point to 7.318 PPL (Table~\ref{tab:l2-mixed-l-ppl}), closing 0.507 PPL or about 55\% of the gap and leaving a residual of 0.421 PPL. Under GPTAQ, the standard \symint\ L=16 point is 7.719 PPL versus a W3A4 GPTAQ reference of 6.509 PPL, a 1.210 PPL gap; replacing \symint\ with \nz lowers the L=16 point to 7.122 PPL (Table~\ref{tab:mixed-l-gptaq-ppl}), closing 0.597 PPL or about 49\% of the gap and leaving a residual of 0.613 PPL. Both iso-method comparisons agree that the level-set redesign alone closes roughly half of the W3A4 gap, while the remaining residual reflects the genuine accuracy cost of keeping half of the transformer layers at two-bit precision rather than using uniform three-bit weights. This mirrors the tail-versus-bulk framing in the introduction: spending a full extra bit of average weight budget to protect sensitive layers still does not reach uniform W3A4, whereas redesigning the four W2 reconstruction levels---at no additional bit cost---recovers roughly half of the same gap. The corresponding mixed-L \wasym point is 7.572, confirming that asymmetric W2 remains a useful diagnostic baseline but not the best mixed-L solution. The qualitative conclusion is stable: the simple L-layer heuristic remains a minor deployment-oriented extension, while the main algorithmic contribution is the fixed no-zero W2 grid redesign.

Table~\ref{tab:l16-downstream} shows downstream accuracy for L=16 and W3A4 references on LLaMA-2-7B.

\begin{table}[t]
\centering
\caption{LLaMA-2-7B KV4 downstream accuracy for W3A4 and L=16 W2A4. \qiftnz and \qiftpot are the proposed grids; Lloyd-Max and NF2 are scalar reference grids. Values are averages over five tasks unless otherwise noted. Higher is better. Bold marks the best non-reference result within each comparison block.}
\label{tab:l16-downstream}
\begin{tabular}{lcccccc}
\toprule
Setting & ARC-C & ARC-E & HellaSwag & PIQA & WinoGrande & Avg \\
\midrule
FP16 & 0.4514 & 0.7382 & 0.7619 & 0.7878 & 0.6938 & 0.6866 \\
W3A4 GPTQ & 0.4002 & 0.6595 & 0.6805 & 0.7345 & 0.6251 & 0.6200 \\
W3A4 GPTAQ & 0.3677 & 0.6448 & 0.6807 & 0.7459 & 0.6504 & 0.6179 \\
L16 sym GPTQ & 0.3584 & 0.6187 & 0.6416 & 0.7252 & 0.6172 & 0.5922 \\
L16 w\_asym GPTQ & 0.3447 & 0.6166 & 0.6441 & 0.7252 & 0.6243 & 0.5910 \\
L16 \qiftnz{} GPTQ & 0.3601 & 0.6317 & 0.6625 & 0.7307 & 0.6385 & 0.6047 \\
L16 \qiftpot{} GPTQ & 0.3763 & 0.6301 & 0.6626 & 0.7399 & 0.6243 & 0.6066 \\
L16 Lloyd-Max GPTQ & 0.3669 & 0.6414 & 0.6679 & 0.7470 & 0.6472 & \textbf{0.6141} \\
L16 NF2 GPTQ & 0.3601 & 0.6212 & 0.6616 & 0.7465 & 0.6164 & 0.6012 \\
\midrule
L16 sym GPTAQ & 0.3626 & 0.6263 & 0.6423 & 0.7242 & 0.6401 & 0.5991 \\
L16 w\_asym GPTAQ & 0.3592 & 0.6313 & 0.6495 & 0.7350 & 0.6275 & 0.6005 \\
L16 \qiftnz{} GPTAQ & 0.3720 & 0.6599 & 0.6712 & 0.7465 & 0.6290 & \textbf{0.6157} \\
L16 \qiftpot{} GPTAQ & 0.3746 & 0.6486 & 0.6650 & 0.7388 & 0.6101 & 0.6074 \\
L16 Lloyd-Max GPTAQ & 0.3592 & 0.6561 & 0.6644 & 0.7416 & 0.6417 & 0.6126 \\
L16 NF2 GPTAQ & 0.3712 & 0.6418 & 0.6622 & 0.7465 & 0.6212 & 0.6086 \\
\bottomrule
\end{tabular}
\end{table}

Although L=16 remains slightly worse than W3A4 in WikiText-2 perplexity under the strict KV4 setting, downstream accuracy shows a much smaller gap. L16 MNZ GPTAQ reaches 0.6157 average accuracy, compared with 0.6179 for W3A4 GPTAQ and 0.6200 for W3A4 GPTQ. Under GPTQ, L16 Lloyd reaches 0.6141, only 0.0059 below W3A4 GPTQ. These results suggest that the simple L-layer heuristic can approach the W3A4 reference on LLaMA-2-7B, while on LLaMA-3.1-8B it still trails W3A4 but substantially improves over the pure W2A4 and standard W2 baselines.

\subsection{Ablation Study}
The RTN bucket diagnostic, ratio sensitivity scan, GPTQ residual analysis, and final PPL/accuracy experiments serve different roles. The bucket diagnostic measures the intrinsic per-output-channel reconstruction behavior of each grid on post-rotation weights. The ratio scan isolates the scale-invariant inner/outer centroid geometry after per-channel normalization using pooled samples and a global scale search. The GPTQ residual analysis checks whether the reconstruction advantage transfers after Hessian-based compensation. Finally, perplexity and downstream accuracy measure the full W2A4/KV4 pipeline. This separation is intentional: RTN diagnostics expose the level-set geometry directly, while GPTQ residuals and final task metrics confirm that the grid advantage survives the full quantization pipeline.

\subsubsection{RTN Bucket Reconstruction Diagnostic}
For each weight sample $w$, we quantize it to a grid level, dequantize it back to $\hat{w}$, and compute squared reconstruction error $(w-\hat{w})^2$. We then aggregate two quantities by assigned bucket:
\begin{align}
    \mathrm{count\%}(g) &= \frac{\#\{w: q(w)=g\}}{\#\{w\}}, \\
    \mathrm{err\%}(g) &= \frac{\sum_{w:q(w)=g}(w-\hat{w})^2}{\sum_w(w-\hat{w})^2}.
\end{align}
This diagnostic is computed on real Hadamard-rotated LLaMA-2-7B weights across all 224 linear modules. It is not a Gaussian simulation and not a downstream metric; it isolates the scalar grid's reconstruction behavior.

This bucket diagnostic uses RTN-style nearest-level quantization rather than GPTQ. For each output channel, we select an MSE-optimized scalar scale by sweeping candidate clipping ratios and then assign each weight to the nearest reconstruction level. Thus, the diagnostic uses per-output-channel scaling, without group-wise quantization, asymmetric zero-points, or Hessian-based GPTQ compensation. The resulting squared error measures the intrinsic reconstruction behavior of the scalar grid on post-rotation weights, rather than the final GPTQ-compensated task error.

\begin{figure}[t]
\centering
\figmaybe{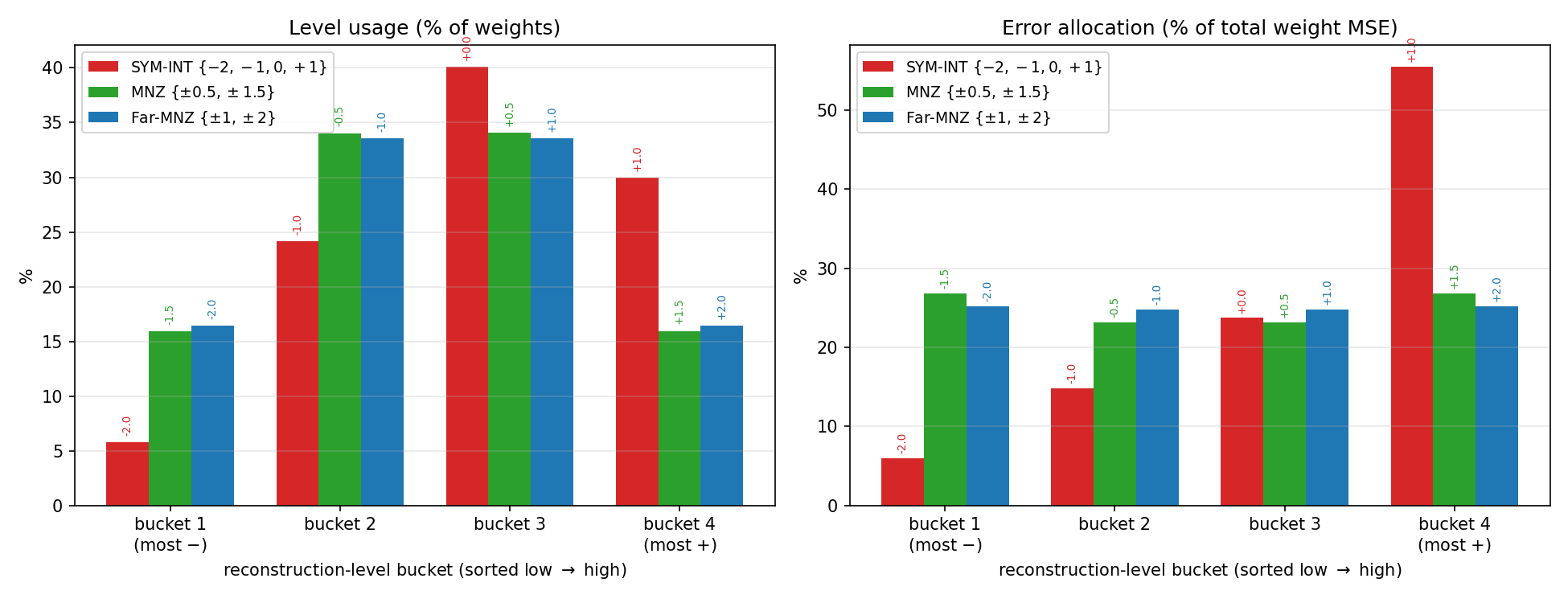}{0.9\linewidth}{Bucket count/error distribution.}
\caption{Bucket count and reconstruction-error allocation for standard W2 and \nz on real rotated LLaMA-2-7B weights. Standard W2 underuses $-2$, overuses zero, and concentrates error in the most positive bucket.}
\label{fig:bucket}
\end{figure}

The asymmetric baseline provides the first indication that modifying reconstruction geometry matters. Bucket-level diagnostics reveal the structural problem in the standard \symint\ grid: Figure~\ref{fig:bucket} shows that the standard W2 grid uses its four levels unevenly. The $-2$ level is rarely used, the zero bucket captures a large fraction of weights, and the $+1$ bucket carries most of the reconstruction error. Importantly, bucket-level error percentages are normalized within each grid and therefore describe where each grid spends its error, not whether the grid has lower total error. Table~\ref{tab:bucket-summary} therefore also reports the aggregate squared reconstruction error over all 6.48B rotated LLaMA-2-7B weights. \nz reduces total squared error from $1.2890\times10^5$ to $1.0228\times10^5$ ($-20.7\%$), and \potwide reduces it to $1.0363\times10^5$ ($-19.6\%$), while also preserving the more balanced no-zero bucket allocation.

\begin{table}[t]
\centering
\caption{Bucket diagnostic summary over real Hadamard-rotated LLaMA-2-7B weights using RTN-style per-output-channel quantization. Bucket error share is normalized within each grid; total squared error is the aggregate nearest-level reconstruction error over all weights. Lower total squared error is better.}
\label{tab:bucket-summary}
\begin{tabular}{llrrrr}
\toprule
Grid & Level & Count (\%) & Bucket err. (\%) & Total sqerr & Rel. total \\
\midrule
\multirow{4}{*}{\symint} & $-2$ & 5.80 & 5.99 & \multirow{4}{*}{$1.2890{\times}10^5$} & \multirow{4}{*}{1.000} \\
 & $-1$ & 24.16 & 14.83 & & \\
 & $0$ & 40.08 & 23.72 & & \\
 & $+1$ & 29.96 & 55.46 & & \\
\midrule
\multirow{4}{*}{\nz} & $-1.5$ & 15.95 & 26.84 & \multirow{4}{*}{$\mathbf{1.0228{\times}10^5}$} & \multirow{4}{*}{\textbf{0.793}} \\
 & $-0.5$ & 34.04 & 23.16 & & \\
 & $+0.5$ & 34.05 & 23.16 & & \\
 & $+1.5$ & 15.96 & 26.84 & & \\
\midrule
\multirow{4}{*}{\potwide} & $-4$ & 17.27 & 29.60 & \multirow{4}{*}{$1.0363{\times}10^5$} & \multirow{4}{*}{0.804} \\
 & $-1$ & 32.72 & 20.40 & & \\
 & $+1$ & 32.73 & 20.42 & & \\
 & $+4$ & 17.27 & 29.59 & & \\
\bottomrule
\end{tabular}
\end{table}

\subsubsection{Where No-Zero Grids Win}
The improvement of \nz is not uniform across all magnitudes: a region-wise error decomposition shows that standard W2 can win very close to zero because of its zero centroid and \farnz loses because its inner centroids sit too far from zero, while \nz wins primarily in the middle high-density region where $\pm0.5$ are closer than either zero or $\pm1$. A zero-bucket counterfactual confirms that \nz is not better on the weights standard W2 snaps to zero, so the total-error reduction in Table~\ref{tab:bucket-summary} is a global centroid-allocation gain rather than a claim that \nz wins on every local region.

\subsubsection{Scale-Invariant Ratio Sensitivity}
\label{sec:ratio-sensitivity}
A previous equal-spacing analysis over $\{-3a,-a,+a,+3a\}$ is useful as a fixed-normalization diagnostic, but if the production quantizer freely optimizes a scale $s$, then
\begin{equation}
    s\{-3a,-a,+a,+3a\} = (sa)\{-3,-1,+1,+3\},
\end{equation}
so $a$ is absorbed into scale and is not a scale-invariant geometry parameter.

We therefore analyze the non-uniform mirror no-zero family
\begin{equation}
    \mathcal{G}(r)=\{-1,-r,+r,+1\}, \quad 0<r<1.
\end{equation}
For each $r$, we search the best global scale $\beta$:
\begin{equation}
    \mathrm{NMSE}(r) = \min_{\beta}\frac{\sum_i\left(x_i - Q_{\beta\mathcal{G}(r)}(x_i)\right)^2}{\sum_i x_i^2}.
\end{equation}
The input samples are real rotated LLaMA-2-7B weights normalized by the standard W2 per-channel scale. After this per-channel normalization, samples from all modules and output channels are pooled. For each ratio $r$, the scan optimizes a single global scale $\beta$ on the pooled samples, rather than re-optimizing a separate scale for every output channel. Therefore, this experiment is a scale-invariant grid-shape diagnostic, not a full production-granularity quantization experiment.

\begin{figure}[t]
\centering
\figmaybe{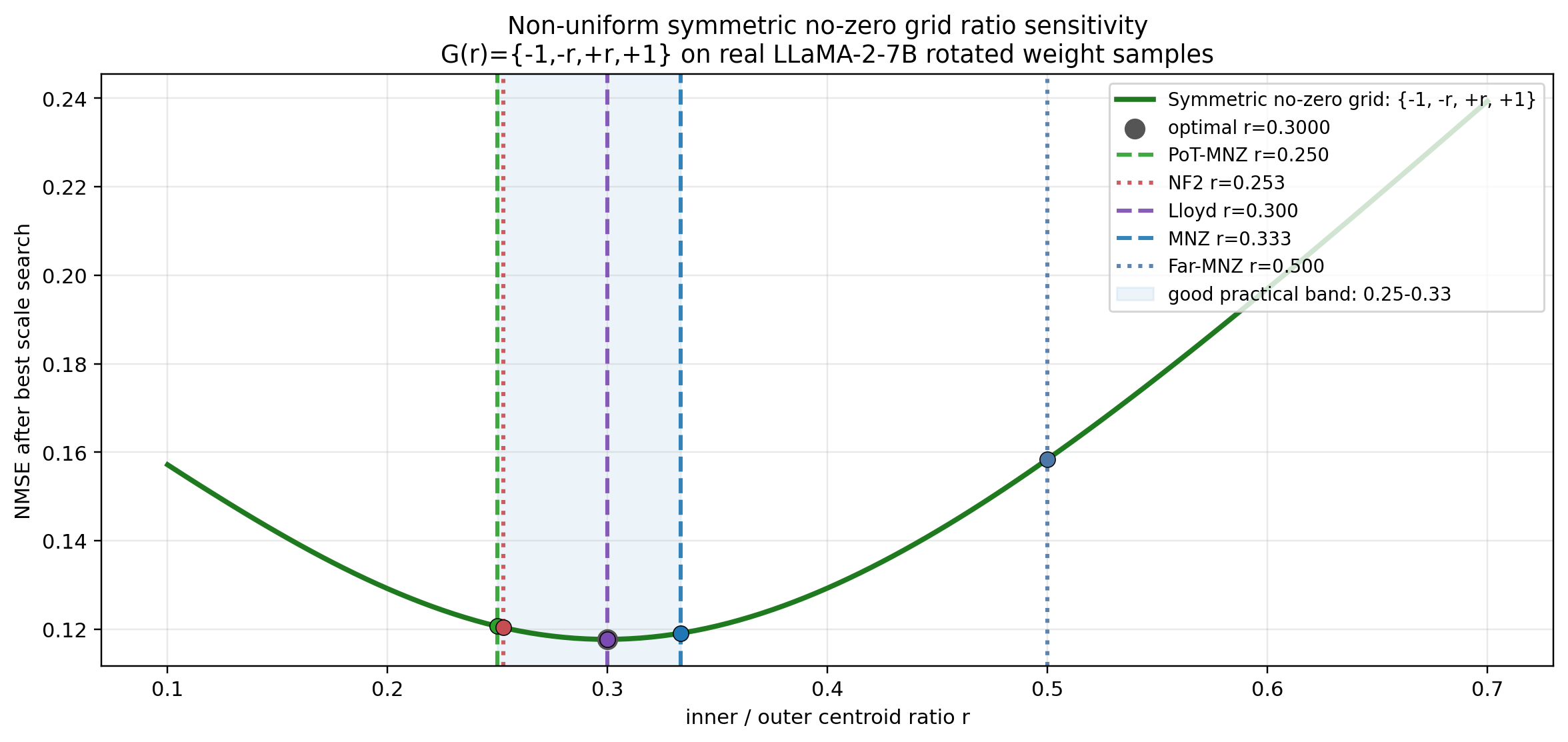}{0.95\linewidth}{Grid ratio sensitivity.}
\caption{Scale-invariant ratio sensitivity over $\{-1,-r,+r,+1\}$ on 100K pooled samples from real rotated LLaMA-2-7B weights after standard W2 per-channel normalization. For each ratio $r$, a single global scale $\beta$ is optimized on the pooled samples before computing NMSE. Good practical grids cluster around $r\approx0.25$--$0.33$, while \farnz at $r=0.5$ places the inner centroids too far from zero.}
\label{fig:ratio-sensitivity}
\end{figure}

Figure~\ref{fig:ratio-sensitivity} sweeps $r$ over the pooled samples and reports the best-scale NMSE for each ratio; reference grids are placed by fixing their corresponding $r$ under the same best-scale search, so the curve isolates the geometry of the four reconstruction levels from arbitrary scale choices. Table~\ref{tab:ratio-sensitivity} lists the exact inner/outer ratio and best-scale NMSE for each grid in this comparison.

\begin{table}[t]
\centering
\caption{Ratio sensitivity reference grids. The discrete scan uses a step of
$0.005$, so its best grid point ($r=0.3000$) can be marginally exceeded by a
reference grid evaluated at its exact ratio (Lloyd-Max, $r=0.2998$); the two are
NMSE-equivalent to within $10^{-6}$, and the best ratio is approximately
$r=0.30$. \qiftnz and \qiftpot are the proposed deployable grids;
Lloyd-Max and NF2 are scalar references; \farnz is a negative diagnostic.}
\label{tab:ratio-sensitivity}
\begin{tabular}{llcc}
\toprule
Role & Grid & Ratio $r$ & NMSE \\
\midrule
Scan-grid optimum & Best ratio on $0.005$ grid & 0.3000 & 0.117719 \\
Reference & Lloyd-Max & 0.2998 & 0.117718 \\
Proposed & \qiftnz & 0.3333 & 0.119115 \\
Reference & \nf & 0.2526 & 0.120367 \\
Proposed & \qiftpot & 0.2500 & 0.120663 \\
Negative diagnostic & \farnz & 0.5000 & 0.158437 \\
\bottomrule
\end{tabular}
\end{table}

NF2 follows the NormalFloat motivation used by QLoRA~\citep{qlora}, while the ratio scan shows that no-zero alone is insufficient: the inner centroids must not be placed too far from zero. Good practical grids have $r\approx0.25$--$0.33$, explaining why \nz, Lloyd, NF2, and \potwide perform well, while \farnz is weaker.

\subsubsection{GPTQ Residual Analysis}
To understand why no-zero grids improve GPTQ results, we record the cumulative residual accumulated during GPTQ quantization. Table~\ref{tab:gptq-residual} shows that the effective no-zero and scalar-reference grids all reduce the cumulative residual relative to standard \symint\ W2. The residual ratio follows the rough PPL trend but does not strictly order the grids by final perplexity: \nz has the lowest residual, whereas Lloyd-Max reaches a slightly lower perplexity.

\begin{table}[t]
\centering
\caption{GPTQ residual accumulation ratio and PPL in a L=16 setting. \qiftnz and \qiftpot are the proposed grids; Lloyd-Max is a scalar reference grid and \farnz is a negative diagnostic. Ratios are normalized by the standard \symint\ grid.}
\label{tab:gptq-residual}
\begin{tabular}{lcc}
\toprule
Grid & PPL & Cumulative residual ratio \\
\midrule
sym & 7.825 & 1.000 \\
\qiftnz & 7.318 & \textbf{0.778} \\
Lloyd-Max & \textbf{7.278} & 0.794 \\
\qiftpot & 7.316 & 0.795 \\
\farnz & 7.982 & 1.020 \\
\bottomrule
\end{tabular}
\end{table}

This analysis does not prove that residual ratio uniquely determines perplexity, but it provides mechanism-level evidence consistent with the overall PPL trend.

\subsection{Summary of Empirical Findings}
Across pure W2A4 perplexity, mixed-L perplexity, downstream accuracy, and GPTQ residual analysis, the consistent trend is that no-zero and Gaussian-aware grids outperform the standard \symint\ W2 grid. Conventional asymmetric W2 improves over the standard grid in many settings, confirming that scalar level-set geometry matters, but it remains worse than source-aware no-zero grids. Lloyd often acts as the strongest scalar reconstruction reference, \nz provides a simple default no-zero grid, and \potwide provides the most hardware-oriented grid while remaining competitive. The L-layer mixed-precision study further shows that the level-set advantage persists under an iso-bit deployment constraint. Overall, the experiments support the central claim that the W2 scalar grid is a major design variable in rotated W2A4/KV4 inference.

\FloatBarrier
\section{Conclusion}

This work argues that W2A4 quantization is not only a bit-width problem
but also a scalar level-set design problem. Treating the four reconstruction
levels of a two-bit weight quantizer as a design choice rather than a
default reveals that the standard \(\{-2,-1,0,+1\}\) grid is
geometrically mismatched to Hadamard-rotated weights, which are
approximately zero-centered and Gaussian-like in standardized shape.
Fixed no-zero grids---\nz\ \(\{\pm0.5,\pm1.5\}\) and \potwide\
\(\{\pm1,\pm4\}\)---match this source by placing two inner centroids
around the dense middle region and two outer centroids in the tails, and
consistently improve perplexity, downstream accuracy, and GPTQ residual
behavior on LLaMA-2-7B and LLaMA-3.1-8B.

\qift is intentionally minimal: its design intervention is confined to the fixed two-bit
code-to-level mapping, leaving rotation, GPTQ/GPTAQ, activation and
KV-cache quantization, and mixed-precision policy unchanged. This makes
it a quantizer-level plug-in component for rotated PTQ pipelines.
Because the decoded levels are small fixed signed integers, \qift is also hardware-friendly: a simple, source-aware step toward accuracy-aware design for extreme low-bit LLM quantization.
This modularity allows it to be combined with future improvements in calibration, rotation, low-bit kernel design, and
hardware-aware mixed-precision policies.

\bibliographystyle{plainnat}

\end{document}